\documentclass[10pt,twocolumn,letterpaper]{article}

\usepackage{iccv}
\usepackage{bm}
\usepackage{times}
\usepackage{epsfig}
\usepackage{graphicx}
\usepackage{amsmath}
\usepackage{amssymb}
\usepackage{microtype}
\usepackage{xcolor}
\usepackage{booktabs}
\usepackage{makecell}
\usepackage{caption}
\usepackage{tabularx}
\usepackage{booktabs}

\iccvfinalcopy
\makeatletter
\usepackage{xspace}
\DeclareRobustCommand\onedot{\futurelet\@let@token\@onedot}
\def\@onedot{\ifx\@let@token.\else.\null\fi\xspace}
\def\eg{\emph{e.g}\onedot}
\def\ie{\emph{i.e}\onedot}
\def\etal{\emph{et~al}\onedot}
\def\etc{\emph{etc}\onedot}
\makeatother

\newif\ifshownotes

\ifdefined\MakeWithNotes
\shownotestrue
\fi
\ifdefined\MakeWithoutNotes
\shownotesfalse
\fi

\ifshownotes
\newcommand{\colornote}[3]{{\color{#1}\bf{#2: #3}\normalfont}}
\newcommand{\colornoteTwo}[3]{{\color{#1}\bf{#3}\normalfont}}
\newcommand{\colornoteThree}[2]{{\color{#1}\bf{#2}\normalfont}}      
\else
\newcommand{\colornote}[3]{}
\newcommand{\colornoteTwo}[3]{}
\newcommand{\colornoteThree}[2]{}      
\fi

\definecolor{darkgreen}{rgb}{0.0,0.65,0}

\def\para#1{\vspace{0.25em}\noindent\textbf{#1}}

\usepackage[pagebackref=true,breaklinks=true,letterpaper=true,colorlinks,bookmarks=false]{hyperref}

\def\iccvPaperID{4487}

\ificcvfinal\fi

\begin{document}
	
	\title{Adding Conditional Control to Text-to-Image Diffusion Models}

\author{Lvmin Zhang, Anyi Rao, and Maneesh Agrawala\\
Stanford University\\
{\tt\small \{lvmin, anyirao, maneesh\}@cs.stanford.edu}
}
	
	\twocolumn[{
		\renewcommand\twocolumn[1][]{#1}
		\vspace{-1em}
		\maketitle
\ificcvfinal\thispagestyle{empty}\fi
		\vspace{-3em}
		\begin{center}
			\centering
			\includegraphics[width=\linewidth]{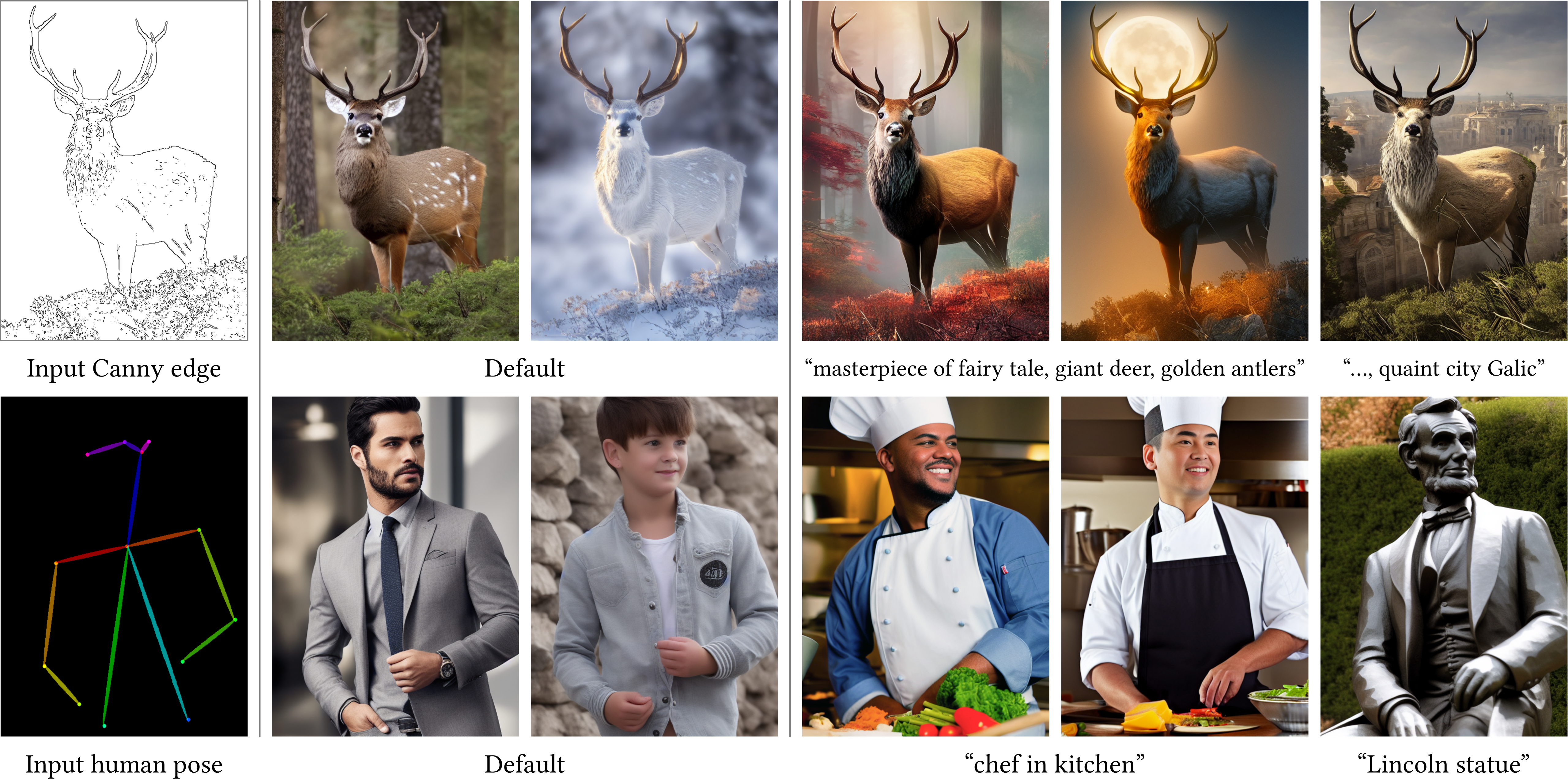}
			\vspace{-2em}
			\captionof{figure}{Controlling Stable Diffusion with learned conditions. ControlNet allows users to add conditions like Canny edges (top), human pose (bottom), \etc, to control the image generation of large pretrained diffusion models. The default results use the prompt ``a high-quality, detailed, and professional image''. Users can optionally give prompts like the ``chef in kitchen''.}
			\label{fig:tea}
		\end{center}
	}]
	
	\begin{abstract}
		We present ControlNet, a neural network architecture to add spatial conditioning controls to large, pretrained text-to-image diffusion models. ControlNet locks the production-ready large diffusion models, and reuses their deep and robust encoding layers pretrained with billions of images as a strong backbone to learn a diverse set of conditional controls. The neural architecture is connected with ``zero convolutions'' (zero-initialized convolution layers) that progressively grow the parameters from zero and ensure that no harmful noise could affect the finetuning. We test various conditioning controls, \eg, edges, depth, segmentation, human pose, \etc, with Stable Diffusion, using single or multiple conditions, with or without prompts. We show that the training of ControlNets is robust with small ($<$50k) and large ($>$1m) datasets. Extensive results show that ControlNet may facilitate wider applications to control image diffusion models.
	\end{abstract}
	
	\section{Introduction}
	\label{sec:intro}
	Many of us have experienced flashes of visual inspiration that
	we wish to capture in a unique image.  With the advent of
	text-to-image diffusion
	models\,\cite{midjourney,DALLE2,rombach2021highresolution}, we
	can now create visually stunning images by typing in a text
	prompt. 
	Yet, text-to-image models are limited in the control
	they provide over the spatial composition of the image;
	precisely expressing complex layouts, poses, shapes and forms
	can be difficult via text prompts alone.
	Generating an image
	that accurately matches our mental imagery often requires numerous
	trial-and-error cycles of editing a prompt, inspecting the
	resulting images and then re-editing the prompt.
	
	Can we enable finer grained spatial control by letting users
	provide additional images that directly specify their desired
	image composition?
	In computer
	vision and machine learning, these additional images (\eg,
	edge maps, human pose skeletons, segmentation maps, depth,
	normals, \etc) are often treated as conditioning on the image
	generation process.  Image-to-image translation
	models\,\cite{isola2017image,CycleGAN2017} learn the mapping
	from conditioning images to target images.  The research
	community has also taken steps to control text-to-image models
	with spatial masks\,\cite{avrahami2022spatext,gafni2022make},
	image editing instructions\,\cite{brooks2022instructpix2pix},
	personalization via
	finetuning\,\cite{gal2022image,ruiz2022dreambooth}, \etc. While
	a few problems (\eg, generating image variations,
	inpainting) can be resolved with training-free techniques like
	constraining the denoising diffusion process or editing
	attention layer activations, a wider variety of problems like
	depth-to-image, pose-to-image, \etc, require end-to-end
	learning and data-driven solutions.
	
	Learning conditional controls for large text-to-image
	diffusion models in an end-to-end way is challenging.  The
	amount of training data for a specific condition may be
	significantly smaller than the data available for general
	text-to-image training. For instance, the largest datasets
	for various specific problems (\eg, object shape/normal, human
	pose extraction, \etc) are usually about 100K in size, which
	is 50,000 times smaller than the
	LAION-5B\,\cite{schuhmann2022laionb} dataset that was used to
	train Stable Diffusion\,\cite{sd15}. The direct finetuning or
	continued training of a large pretrained model with limited
	data may cause overfitting and catastrophic
	forgetting\,\cite{hu2021lora,ruiz2022dreambooth}.
	Researchers have shown that such forgetting can be alleviated by restricting the number or rank of trainable parameters\,\cite{chen2022vision,ha2017hypernetworks,hu2021lora,zhang2020side}.
	For our problem, designing deeper or more customized neural architectures might be necessary for handling in-the-wild conditioning images with complex shapes and diverse high-level semantics.
	
	This paper presents ControlNet, an end-to-end neural
	network architecture that learns conditional controls for
	large pretrained text-to-image diffusion models (Stable
	Diffusion in our implementation).  ControlNet preserves the
	quality and capabilities of the large model by locking its
	parameters, and also making a {\em trainable copy} of its
	encoding layers. This architecture treats the large pretrained
	model as a strong backbone for learning diverse conditional
	controls.  The trainable copy and the original, locked model
	are connected with {\em zero convolution} layers, with weights
	initialized to zeros so that they progressively grow during
	the training. This architecture ensures that harmful noise is
	not added to the deep features of the large diffusion model at
	the beginning of training, and protects the large-scale
	pretrained backbone in the trainable copy from being damaged
	by such noise.

	Our experiments show that ControlNet can control Stable
	Diffusion with various conditioning inputs, including Canny
	edges, Hough lines, user scribbles, human key points,
	segmentation maps, shape normals, depths,
	\etc. (Figure\,\ref{fig:tea}).  We test our approach using a
	single conditioning image, with or without text prompts, and
	we demonstrate how our approach supports the composition of
	multiple conditions.  Additionally, we report that the
	training of ControlNet is robust and scalable on datasets of
	different sizes, and that for some tasks like depth-to-image
	conditioning, training ControlNets on a single NVIDIA RTX
	3090Ti GPU can achieve results competitive with industrial
	models trained on large computation clusters.  Finally, we
	conduct ablative studies to investigate the contribution of
	each component of our model, and compare our models to several
	strong conditional image generation baselines with user
	studies.
	
	In summary, (1) we propose ControlNet, a neural network
	architecture that can add spatially localized input conditions
	to a pretrained text-to-image diffusion model via efficient
	finetuning, (2) we present pretrained ControlNets to control
	Stable Diffusion, conditioned on Canny edges, Hough lines,
	user scribbles, human key points, segmentation maps, shape
	normals, depths, and cartoon line drawings, and (3) we
	validate the method with ablative experiments comparing to
	several alternative architectures, and conduct user studies
	focused on several previous baselines across different tasks.
	
	\section{Related Work}
	\label{sec:related}
	
	\subsection{Finetuning Neural Networks}
	
	\noindent One way to finetune a neural network is to directly continue training
	it with the additional training data. But this approach can lead to
	overfitting, mode collapse, and catastrophic forgetting.  Extensive
	research has focused on developing finetuning strategies that avoid
	such issues.
	
	\para{HyperNetwork} is an approach that originated in the Natural
	Language Processing (NLP) community\,\cite{ha2017hypernetworks}, with
	the aim of training a small recurrent neural network to influence the
	weights of a larger one.  It has been applied to image generation with
	generative adversarial networks (GANs)\,\cite{alaluf2022hyperstyle,
		dinh2022hyperinverter}.  Heathen~\etal\,\cite{heathen} and Kurumuz\,\cite{nai} implement
	HyperNetworks for Stable Diffusion\,\cite{rombach2021highresolution}
	to change the artistic style of its output images.
	
	\para{Adapter} methods are widely used in NLP for customizing a
	pretrained transformer model to other tasks by embedding new module
	layers into it\,\cite{houlsby2019parameter,stickland2019bert}.  In
	computer vision, adapters are used for incremental
	learning\,\cite{rosenfeld2018incremental} and domain
	adaptation\,\cite{rebuffi2018efficient}.  This technique is often used
	with CLIP\,\cite{radford2021learning} for transferring pretrained
	backbone models to different
	tasks\,\cite{gao2021clip,radford2021learning,sung2021vl,zhang2021tip}.
	More recently, adapters have yielded successful results in vision
	transformers\,\cite{li2022exploring,li2021benchmarking} and
	ViT-Adapter\,\cite{chen2022vision}.  In concurrent work with ours,
	T2I-Adapter\,\cite{mou2023t2i} adapts Stable Diffusion to external
	conditions. 
	
	\para{Additive Learning} circumvents forgetting by freezing the
	original model weights and adding a small number of new parameters using learned weight
	masks\,\cite{mallya2018piggyback,rosenfeld2018incremental},
	pruning\,\cite{mallya2018packnet}, or hard
	attention\,\cite{serra2018overcoming}.
	Side-Tuning\,\cite{zhang2020side} uses a side branch model to learn
	extra functionality by linearly blending the outputs of a frozen model
	and an added network, with a predefined blending weight schedule.
	
	\para{Low-Rank Adaptation (LoRA)} prevents catastrophic
	forgetting\,\cite{hu2021lora} by learning the offset of parameters with low-rank matrices, based on the observation that many
	over-parameterized models reside in a low intrinsic
	dimension subspace\,\cite{aghajanyan2021intrinsic,li2018measuring}.
	
	\para{Zero-Initialized Layers} are used by ControlNet for connecting
	network blocks. Research on neural networks has extensively discussed
	the initialization and manipulation of network
	weights\,\cite{karras2017progressive,karras2019style,LeCun2015,lecun1998gradient,lehtinen2018noise2noise,Rumelhart1986,sdd,zhao2021zero}.
	For example, Gaussian initialization of weights can be less risky than
	initializing with zeros\,\cite{arfin2020weight}.  More recently,
	Nichol~\etal~\cite{nichol2021improved} discussed how to scale the
	initial weight of convolution layers in a diffusion model to improve
	the training, and their implementation of ``zero\_module'' is an
	extreme case to scale weights to zero.  Stability's model
	cards\,\cite{sdd} also mention the use of zero weights in neural
	layers.  Manipulating the initial convolution weights is also
	discussed in ProGAN\,\cite{karras2017progressive},
	StyleGAN\,\cite{karras2019style},
	and Noise2Noise\,\cite{lehtinen2018noise2noise}.
	
	\subsection{Image Diffusion}
	
	\para{Image Diffusion Models} were first introduced by
	Sohl-Dickstein~\etal\,\cite{sohl2015deep} and have been recently
	applied to image
	generation\,\cite{dhariwal2021diffusion,kingma2021variational}.
	The Latent Diffusion Models (LDM)\,\cite{rombach2021highresolution}
	performs the diffusion steps in the latent image
	space\,\cite{esser2021taming}, which reduces the computation cost.
	Text-to-image diffusion
	models achieve state-of-the-art image generation results by encoding
	text inputs into latent vectors via pretrained language models like
	CLIP\,\cite{radford2021learning}.  Glide\,\cite{nichol2021glide} is a
	text-guided diffusion model supporting image generation and editing.
	Disco Diffusion\,\cite{disco} processes text prompts with clip guidance.
	Stable Diffusion\,\cite{sd15} is a large-scale implementation of
	latent diffusion\,\cite{rombach2021highresolution}.
	Imagen\,\cite{saharia2022photorealistic} directly diffuses pixels
	using a pyramid structure without using latent images.  Commercial
	products include DALL-E2\,\cite{DALLE2} and
	Midjourney\,\cite{midjourney}.
	
	\para{Controlling Image Diffusion Models} facilitate personalization,
	customization, or task-specific image generation.  The image diffusion
	process directly provides some control over color
	variation\,\cite{meng2021sdedit} and
	inpainting\,\cite{ramesh2022hierarchical,avrahami2022blended}.
	Text-guided control methods focus on adjusting prompts, manipulating
	CLIP features, and modifying
	cross-attention\,\cite{avrahami2022blended, brooks2022instructpix2pix,
		gafni2022make,hertz2022prompt, kawar2022imagic,kim2022diffusionclip,
		nichol2021glide,parmar2023zero,ramesh2022hierarchical}.
	MakeAScene\,\cite{gafni2022make} encodes segmentation masks into
	tokens to control image generation.
	SpaText\,\cite{avrahami2022spatext} maps segmentation masks into
	localized token embeddings.  GLIGEN\,\cite{li2023gligen} learns new
	parameters in attention layers of diffusion models for grounded
	generating.  Textual Inversion\,\cite{gal2022image} and
	DreamBooth\,\cite{ruiz2022dreambooth} can personalize content in the
	generated image by finetuning the image diffusion model using a small
	set of user-provided example images.  Prompt-based image
	editing\,\cite{brooks2022instructpix2pix,huang2023region,pnpDiffusion2022}
	provides practical tools to manipulate images with prompts.
	Voynov\,\etal\,\cite{voynov2022sketch} propose an optimization method
	that fits the diffusion process with sketches.  Concurrent works
	\cite{bar2023multidiffusion,bashkirova2023masksketch,lhhuang2023composer,mou2023t2i}
	examine a wide variety of ways to control diffusion models.
	
	\subsection{Image-to-Image Translation}
	
	\noindent Conditional
	GANs\,\cite{choi2018stargan,isola2017image,park2019semantic,wang2018high,zhang2020cross,zhou2021cocosnet,CycleGAN2017,zhu2017toward} and transformers\,\cite{chen2021pre,esser2021taming, ramesh2021zero} can learn the mapping between different image domains, \eg, Taming
	Transformer\,\cite{esser2021taming} is a vision transformer
	approach; Palette\,\cite{saharia2022palette} is a conditional
	diffusion model trained from scratch; PITI\,\cite{wang2022pretraining} is a pretraining-based conditional
	diffusion model for image-to-image translation. Manipulating pretrained GANs can handle
	specific image-to-image tasks, \eg, StyleGANs can be
	controlled by extra encoders \cite{richardson2021encoding}, with more
	applications studied in
	\cite{alaluf2021matter,gal2022stylegan,karras2021style,katzir2022multi,mokady2022selfdistilled,nitzan2022mystyle,Patashnik_2021_ICCV,richardson2021encoding}.
	
	\section{Method}
	\label{sec:method}
	
	\begin{figure}
		\vspace{-10pt}
		\includegraphics[width=\linewidth]{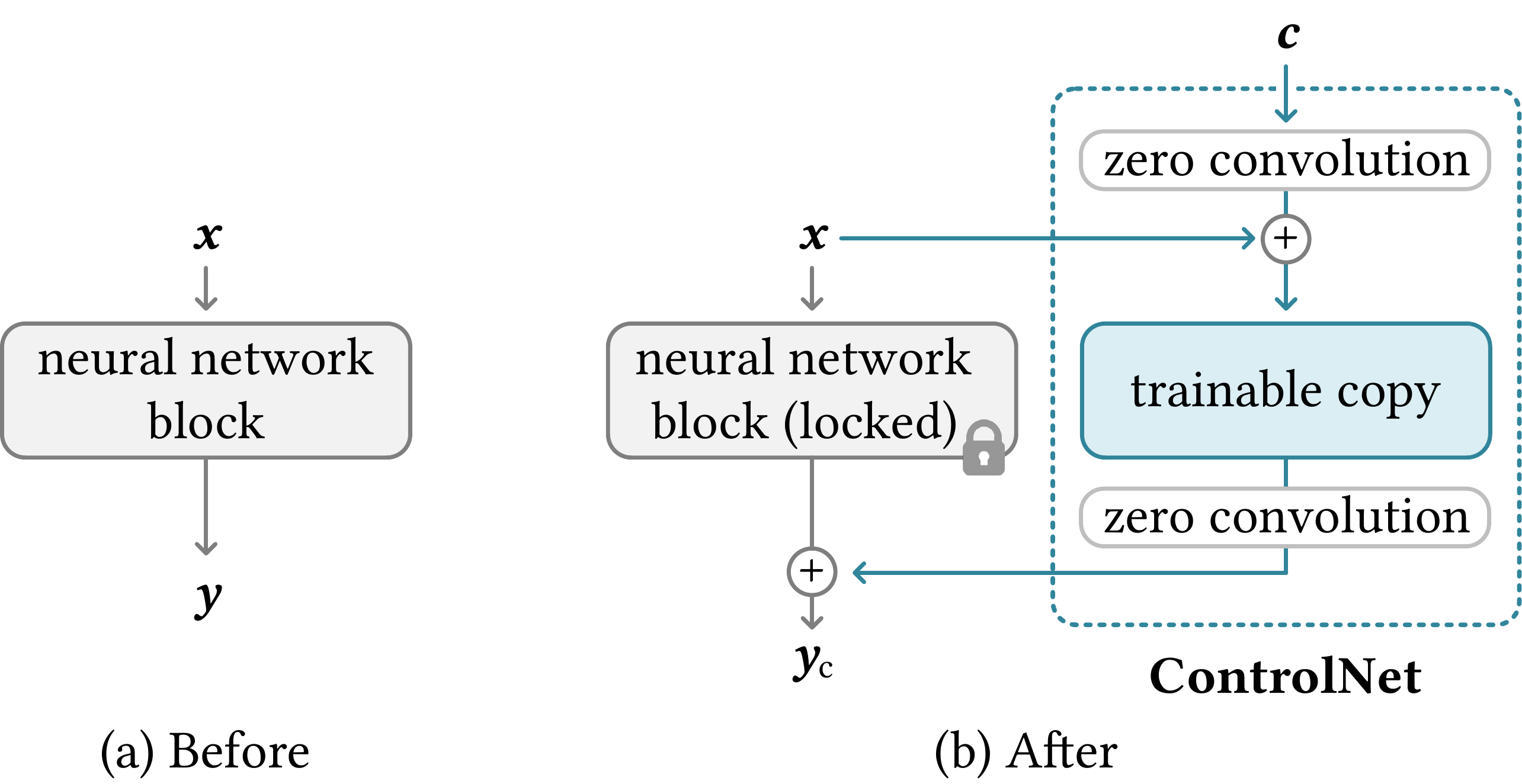}
		\vspace{-17pt}
		\caption{A neural block takes a feature map $x$ as input
			and outputs another feature map $y$, as shown in (a). To add a ControlNet to
			such a block we lock the original block and create a trainable copy 
			and connect them together using zero
			convolution layers, \ie, $1\times 1$ convolution with
			both weight and bias initialized to zero. Here $c$ is a conditioning vector
			that we wish to add to the network, as shown in (b).}
		\vspace{-7pt}
		\label{fig:he}
	\end{figure}
	
	ControlNet is a neural network architecture that can enhance large
	pretrained text-to-image diffusion models with spatially localized,
	task-specific image conditions.  We first introduce the basic
	structure of a ControlNet in Section\,\ref{sec:control} and then
	describe how we apply a ControlNet to the image diffusion model Stable
	Diffusion\,\cite{rombach2021highresolution} in
	Section\,\ref{sec:controldiff}.  We elaborate on our training in
	Section~\ref{sec:train} and detail several extra considerations during
	inference such as composing multiple ControlNets in
	Section\,\ref{sec:infer}.
	
	\subsection{ControlNet}
	\label{sec:control}
	
	ControlNet injects additional conditions into the blocks of a neural
	network (Figure\,\ref{fig:he}). Herein, we use the term {\em
		network block} to refer to a set of neural layers that are commonly
	put together to form a single unit of a neural network, \eg, resnet block,
	conv-bn-relu block, multi-head attention block, transformer
	block,\,\etc.
	Suppose $\mathcal{F}(\cdot;\Theta)$ is such a trained neural block, with
	parameters $\Theta$, that transforms an input feature map $\bm{x}$,
	into another feature map $\bm{y}$ as
	\vspace{-3pt}\begin{equation}\vspace{-3pt}
		\bm{y}=\mathcal{F}(\bm{x};\Theta) .
	\end{equation}
	In our setting, $\bm{x}$ and $\bm{y}$ are usually 2D feature maps, \ie, $\bm{x}\in\mathbb{R}^{h\times w \times c}$ with $\{h, w, c\}$
	as the height, width, and number of channels in the map,
	respectively (Figure\,\ref{fig:he}a). 
	
	To add a ControlNet to such a pre-trained neural block, we lock (freeze)
	the parameters $\Theta$ of the original block and simultaneously clone
	the block to a {\em trainable copy} with parameters $\Theta_\text{c}$
	(Figure\,\ref{fig:he}b).  The trainable copy takes an external
	conditioning vector $\bm{c}$ as input. When this structure is applied
	to large models like Stable Diffusion, the locked parameters preserve
	the production-ready model trained with billions of images, while the
	trainable copy reuses such large-scale pretrained model to establish a
	deep, robust, and strong backbone for handling diverse input
	conditions.
	
	The trainable copy is connected to the locked model with {\em zero convolution} layers, denoted 
	$\mathcal{Z}(\cdot;\cdot)$. Specifically,
	$\mathcal{Z}(\cdot;\cdot)$ is a $1\times 1$ convolution layer with
	both weight and bias initialized to zeros. 
	To build up a ControlNet, we use two instances of zero convolutions with parameters $\Theta_\text{z1}$
	and $\Theta_\text{z2}$ respectively. The complete ControlNet then
	computes
	\vspace{-3pt}\begin{equation}\vspace{-3pt}
		\label{eq:key1}
		\bm{y}_\text{c}=\mathcal{F}(\bm{x};\Theta)+\mathcal{Z}(\mathcal{F}(\bm{x}+\mathcal{Z}(\bm{c};\Theta_\text{z1});\Theta_\text{c});\Theta_\text{z2}),
	\end{equation}
	where $\bm{y}_\text{c}$ is the output of the ControlNet block.
	In the first training step, since both the weight and bias parameters of a zero convolution layer are initialized to zero, both of the $\mathcal{Z}(\cdot;\cdot)$ terms in Equation\,\eqref{eq:key1} evaluate to zero, 
	and
	\vspace{-3pt}\begin{equation}\vspace{-3pt}
		\label{key3}
		\bm{y}_\text{c} = \bm{y}.
	\end{equation}
	In this way, harmful noise cannot influence the hidden states of
	the neural network layers in the trainable copy when the training starts.
	Moreover,
	since $\mathcal{Z}(\bm{c};\Theta_\text{z1})=\bm{0}$ and the trainable
	copy also receives the input image $\bm{x}$, the
	trainable copy is fully functional and retains the capabilities of the large, pretrained model
	allowing it to serve as a strong backbone for further
	learning. Zero convolutions protect this backbone by
	eliminating random noise as gradients in the initial
	training steps. We detail the gradient calculation for zero
	convolutions in supplementary materials.
	
	\begin{figure}
		\includegraphics[width=\linewidth]{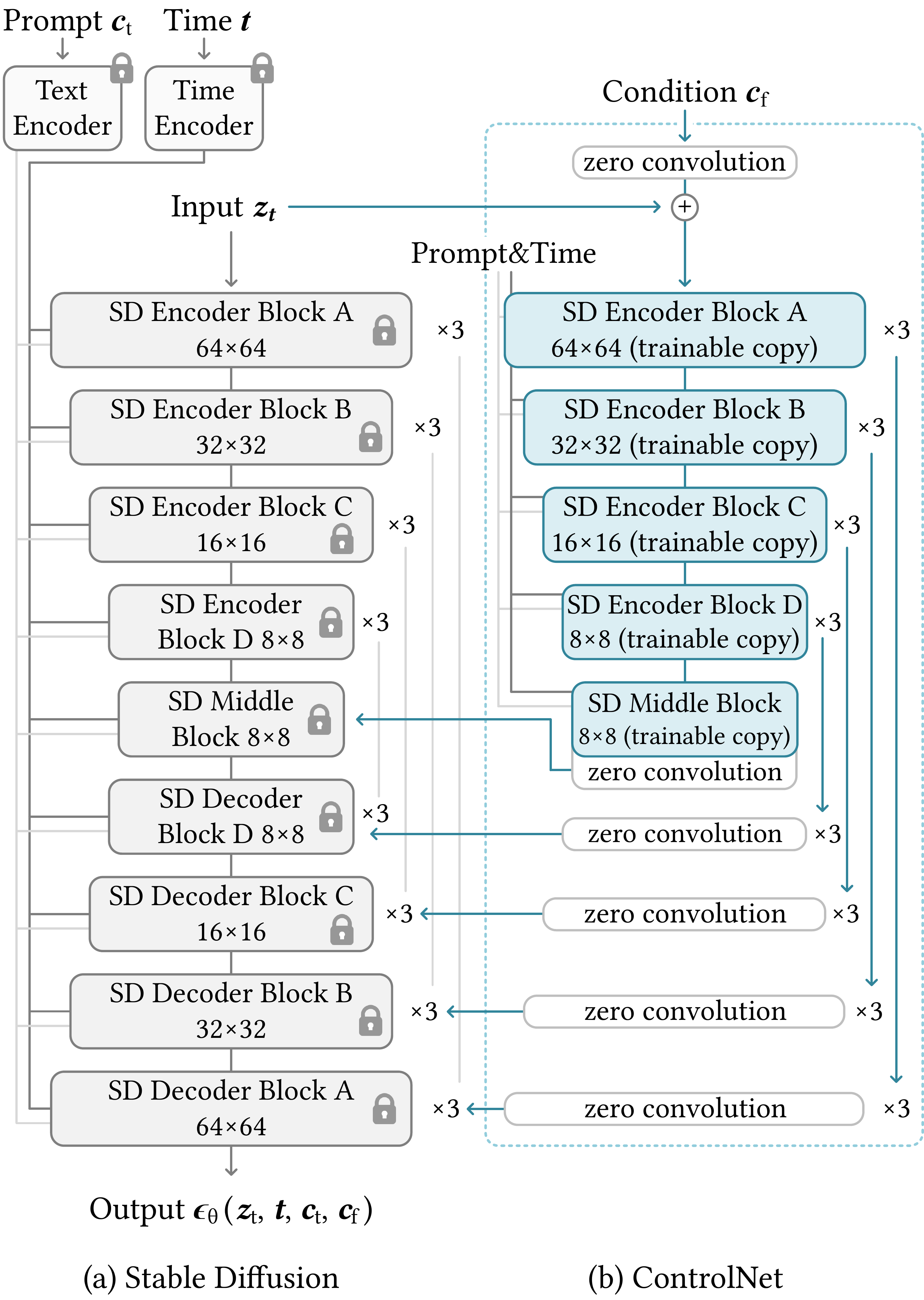}
		\vspace{-17pt}
		\caption{Stable Diffusion's U-net architecture connected with a ControlNet on the encoder blocks and middle block. The locked, gray blocks
			show the structure of Stable Diffusion V1.5 (or V2.1, as they use the same U-net architecture). The trainable blue blocks and the white zero convolution layers are added to build a ControlNet.}
		\vspace{-7pt}
		\label{fig:hesd} 
	\end{figure}
	
	\subsection{ControlNet for Text-to-Image Diffusion}
	\label{sec:controldiff}
	
	We use Stable Diffusion\,\cite{rombach2021highresolution} as an
	example to show how ControlNet can add conditional control to a large
	pretrained diffusion model.  Stable Diffusion is essentially a
	U-Net\,\cite{ronneberger2015u} with an encoder, a middle block, and a
	skip-connected decoder. Both the encoder and decoder contain 12
	blocks, and the full model contains 25 blocks, including the middle
	block. Of the 25 blocks, 8 blocks are down-sampling or up-sampling
	convolution layers, while the other 17 blocks are main blocks that
	each contain 4 resnet layers and 2 Vision Transformers (ViTs).  Each
	ViT contains several cross-attention and self-attention mechanisms.
	For example, in Figure\,\ref{fig:hesd}a, the
	``SD Encoder Block A'' contains 4 resnet layers and 2 ViTs, while the 
	``$\times 3$'' indicates that this block is repeated three times.
	Text prompts are encoded using the CLIP text encoder\,\cite{radford2021learning}, and diffusion timesteps are encoded
	with a time encoder using positional encoding.
	
	The ControlNet structure is applied to each encoder level of the U-net
	(Figure\,\ref{fig:hesd}b).  In particular, we use ControlNet to create a
	trainable copy of the 12 encoding blocks and 1 middle block of Stable
	Diffusion. The 12 encoding blocks are in 4 resolutions
	($64\times64,32\times32,16\times16,8\times8$) with each one replicated 3
	times.  The outputs are added to the 12 skip-connections and 1 middle
	block of the U-net.  Since Stable Diffusion is a typical U-net structure, this
	ControlNet architecture is likely to be applicable with other
	models.
	
	The way we connect the ControlNet is computationally efficient ---
	since the locked copy parameters are frozen, no gradient computation
	is required in the originally locked encoder for the finetuning.  This
	approach speeds up training and saves GPU memory. As tested on a
	single NVIDIA A100 PCIE 40GB, optimizing Stable Diffusion with
	ControlNet requires only about 23\% more GPU memory and 34\% more time
	in each training iteration, compared to optimizing Stable Diffusion
	without ControlNet.
	
	Image diffusion models learn to progressively denoise images and
	generate samples from the training domain.  The denoising process can
	occur in pixel space or in a {\em latent} space encoded from training
	data. Stable Diffusion uses latent images as the training domain as
	working in this space has been shown to stabilize the
	training process\,\cite{rombach2021highresolution}.
	Specifically, Stable Diffusion uses a pre-processing method similar to
	VQ-GAN~\cite{esser2021taming} to convert $512\times 512$ pixel-space images into
	smaller $64\times 64$ {\em latent images}.
	To add ControlNet to Stable Diffusion, 
	we first convert each input conditioning image (\eg, edge, pose, depth, \etc) from an input size of $512\times 512$ into a
	$64\times 64$ feature space vector that matches the size of Stable Diffusion.  
	In particular, we use a tiny network
	$\mathcal{E}(\cdot)$ of four convolution layers with $4\times 4$
	kernels and $2 \times 2$ strides (activated by ReLU, using 16, 32, 64,
	128, channels respectively, initialized with Gaussian weights and
	trained jointly with the full model) to encode an image-space
	condition $\bm{c}_\text{i}$ into a feature space conditioning vector
	$\bm{c}_\text{f}$ as,
	\vspace{-3pt}\begin{equation}\vspace{-3pt}
		\bm{c}_\text{f}=\mathcal{E}(\bm{c}_\text{i}).
	\end{equation}
	The conditioning vector $\bm{c}_\text{f}$ is passed into the ControlNet.
	
	\subsection{Training}
	\label{sec:train}
	
	Given an input image $\bm{z}_0$, image diffusion algorithms
	progressively add noise to the image and produce a noisy image
	$\bm{z}_t$, where $t$ represents the number of times noise is added. Given a
	set of conditions including time step $\bm{t}$, text prompts $\bm{c}_t$, as
	well as a task-specific condition $\bm{c}_\text{f}$, image diffusion
	algorithms learn a network $\epsilon_\theta$ to predict the noise
	added to the noisy image $\bm{z}_t$ with 
	\vspace{-3pt}\begin{equation}\vspace{-3pt}
		\mathcal{L} = \mathbb{E}_{\bm{z}_0, \bm{t}, \bm{c}_t, \bm{c}_\text{f}, \epsilon \sim \mathcal{N}(0, 1) }\Big[ \Vert \epsilon - \epsilon_\theta(\bm{z}_{t}, \bm{t}, \bm{c}_t, \bm{c}_\text{f})) \Vert_{2}^{2}\Big],
		\label{eq:loss}
	\end{equation}
	where $\mathcal{L}$ is the overall learning objective of the entire diffusion model.
	This learning objective is directly used in finetuning diffusion models with ControlNet.
	
	In the training process, we randomly replace 50\% text prompts
	$\bm{c}_t$ with empty strings. 
	This approach increases ControlNet's
	ability to directly recognize semantics in the input conditioning images
	(\eg, edges, poses, depth, \etc) as a replacement for the prompt.
	
	\begin{figure}[!t]
		\includegraphics[width=\linewidth]{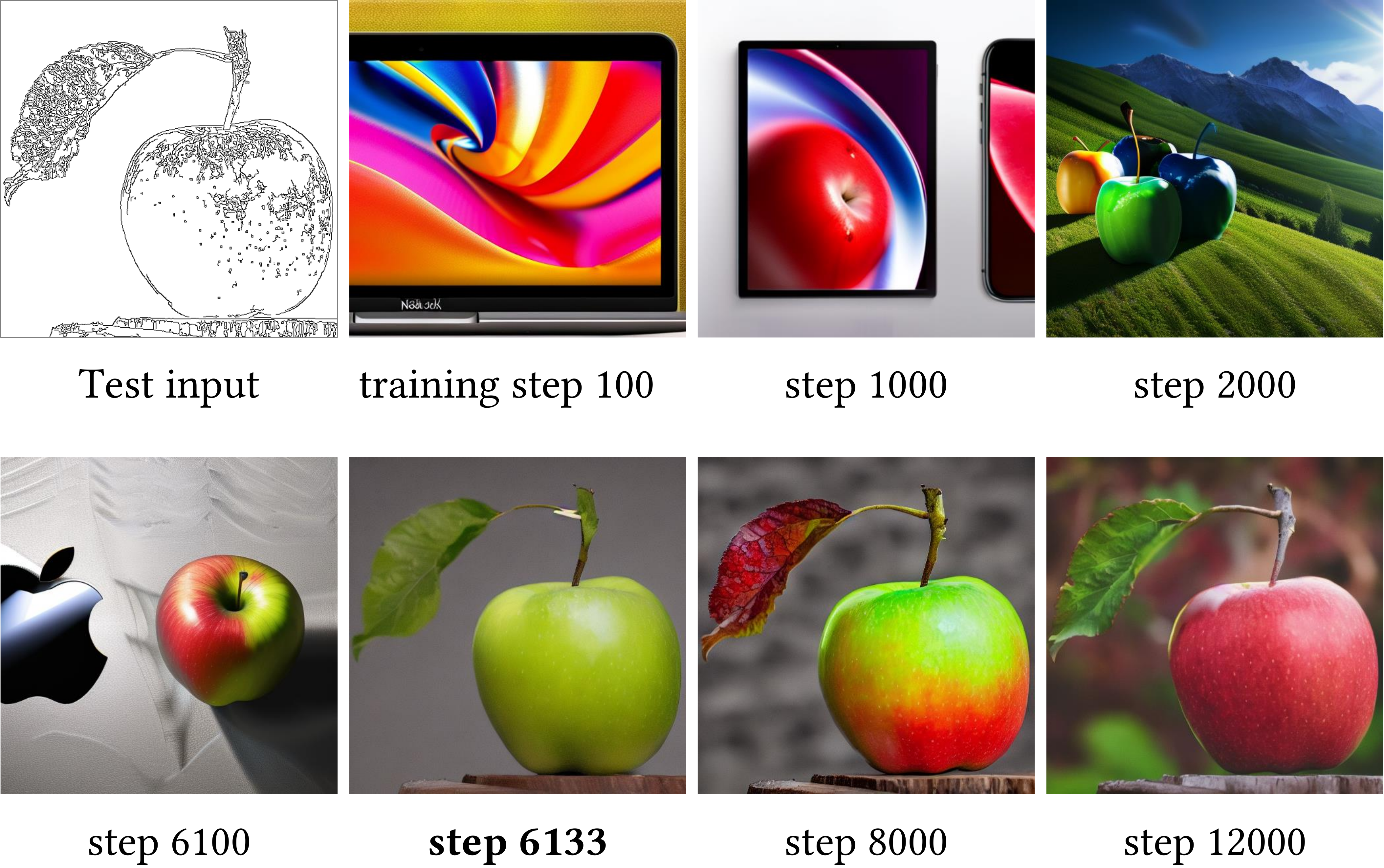}
		\vspace{-17pt}
		\caption{The sudden convergence phenomenon. 
			Due to the zero convolutions, ControlNet
			always predicts high-quality images during the
			entire training.  At a certain step in the
			training process (\eg, the 6133 steps marked
			in bold), the model suddenly learns to follow
			the input condition.}
		\vspace{-7pt}
		\label{fig:train}
	\end{figure}
	
	\begin{figure}[!t]
		\includegraphics[width=\linewidth]{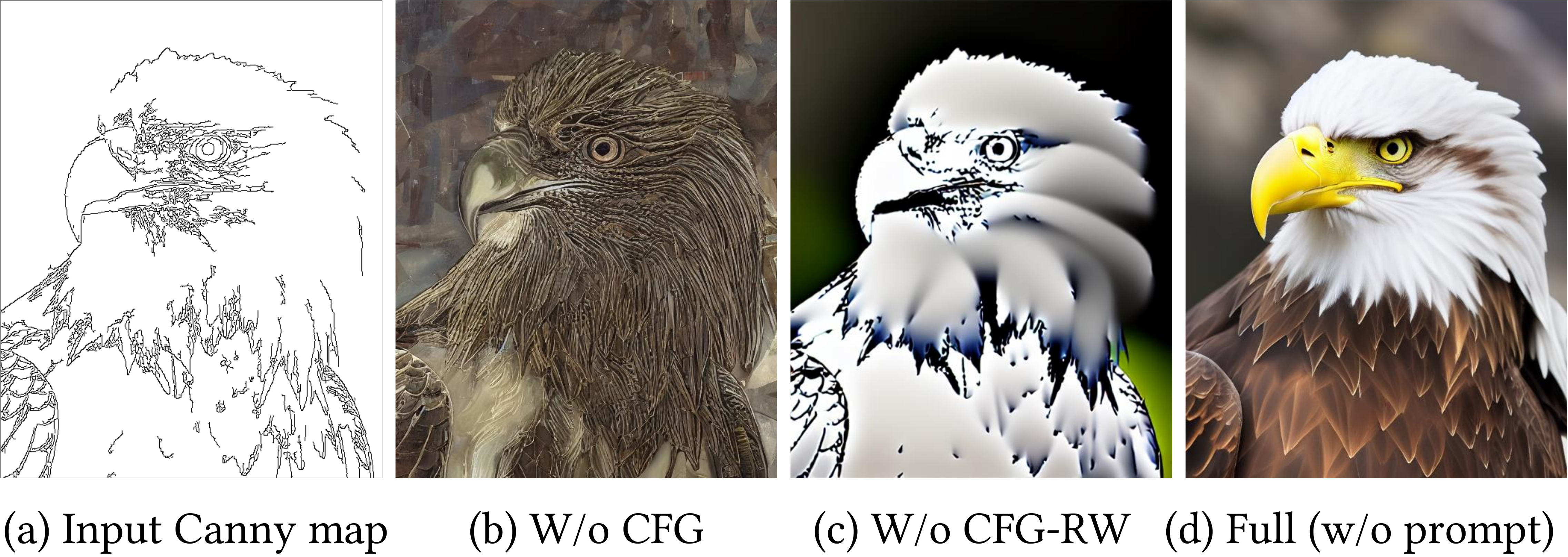}
		\vspace{-17pt}
		\caption{Effect of Classifier-Free Guidance (CFG) and the proposed CFG Resolution Weighting (CFG-RW).}
		\vspace{-7pt}
		\label{fig:cfg}
	\end{figure}
	
	\begin{figure}[!t]
		\includegraphics[width=\linewidth]{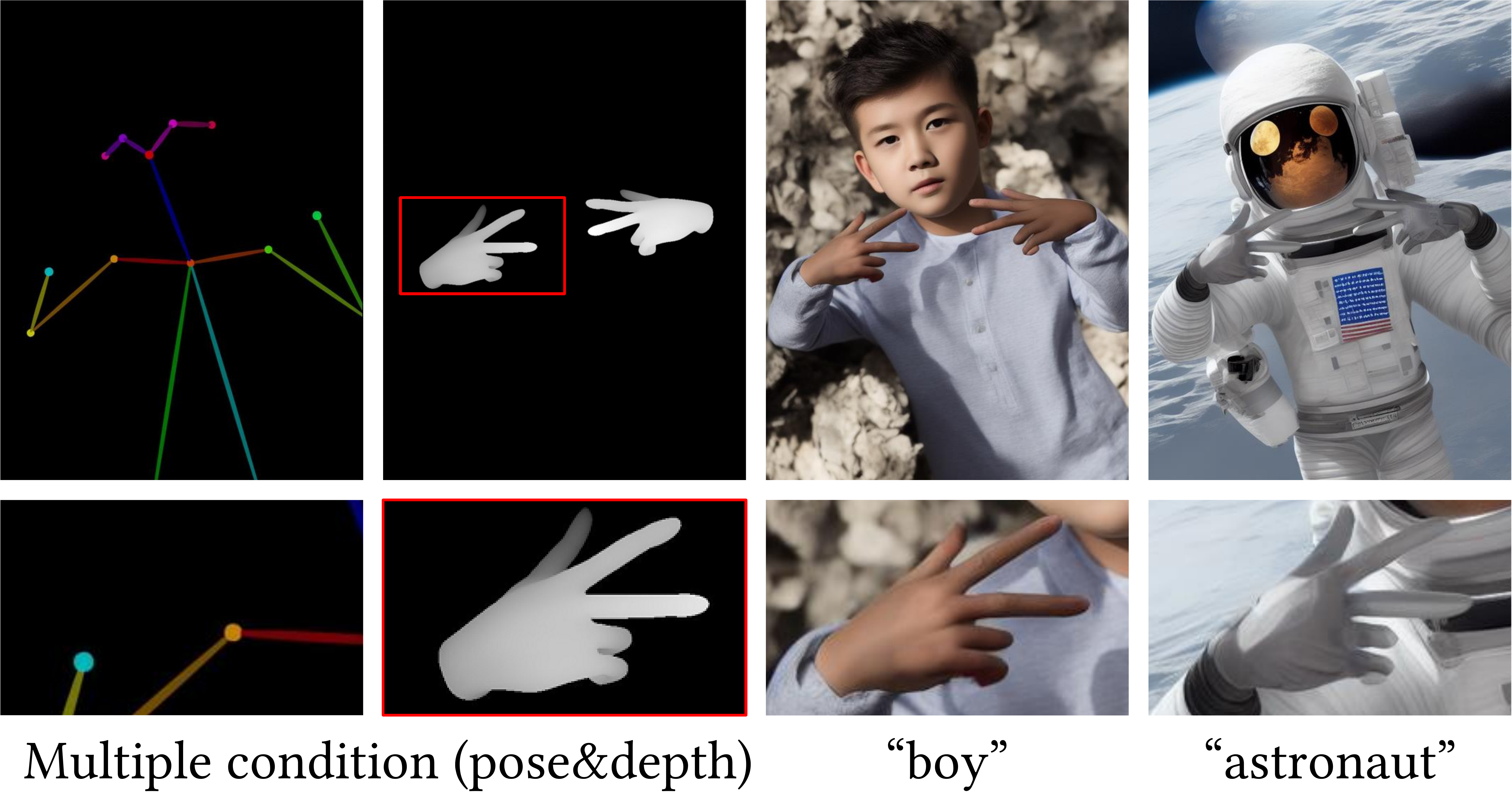}
		\vspace{-22pt}
		\caption{Composition of multiple conditions. We present the application to use depth and pose simultaneously.}
		\label{fig:multi}
		\vspace{-10pt}
	\end{figure}
	
	During the training process, since zero convolutions do not add noise to the network, the model should always be able to predict high-quality images. We observe that the model does not gradually learn the control conditions but abruptly succeeds in following the input conditioning image; usually in less than 10K optimization steps. As shown in Figure\,\ref{fig:train}, we call this the ``sudden convergence phenomenon''.
	
	\begin{figure*}
		\centering
		\resizebox{\textwidth}{!}{
			\begin{tabularx}{1.28\textwidth}{*{8}{>{\centering\arraybackslash}X}}
				Sketch & 
				Normal map & 
				Depth map & 
				Canny\cite{canny1986computational} edge & 
				M-LSD\cite{gu2021realtime} line& 
				HED\cite{xie2015holistically} edge& 
				ADE20k\cite{zhou2017scene}~seg.& 
				Human pose
		\end{tabularx}}
		\includegraphics[width=\linewidth]{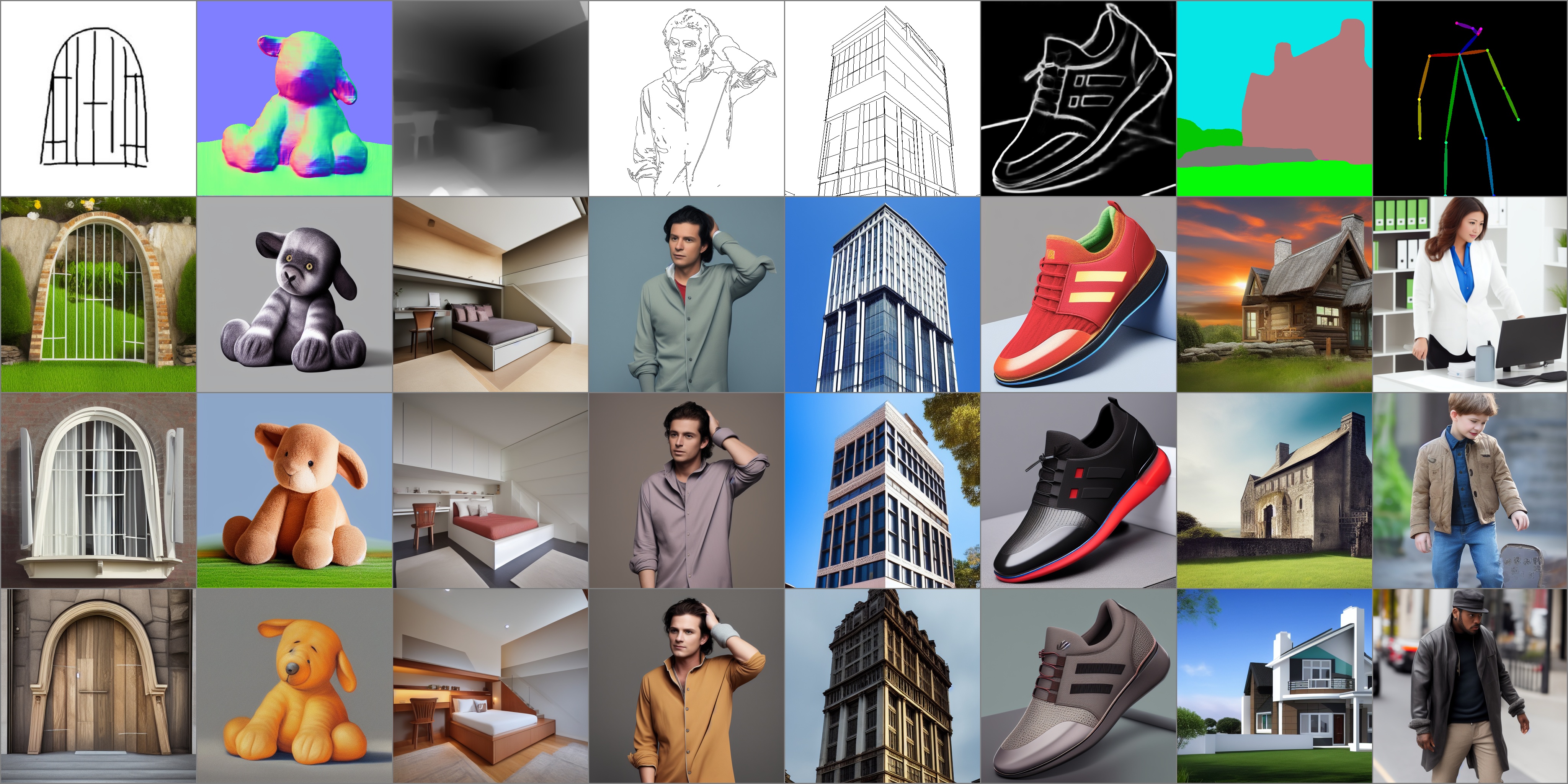}
		\vspace{-17pt}
		\caption{Controlling Stable Diffusion with various conditions \textbf{without prompts}. The top row is input conditions, while all other rows are outputs. We use the empty string as input prompts. All models are trained with general-domain data. The model has to recognize semantic contents in the input condition images to generate images.}
		\vspace{-7pt}
		\label{fig:qua}
	\end{figure*}
	
	\subsection{Inference}
	\label{sec:infer}
	
	We can further control how the extra conditions of ControlNet affect the denoising diffusion process
	in several ways.
	
	
	\para{Classifier-free guidance resolution weighting.} Stable Diffusion
	depends on a technique called Classifier-Free Guidance
	(CFG)\,\cite{cfg} to generate high-quality images. CFG is
	formulated as
	$\epsilon_\text{prd}=\epsilon_\text{uc}+\beta_\text{cfg}(\epsilon_\text{c}-\epsilon_\text{uc})$
	where
	$\epsilon_\text{prd}$, $\epsilon_\text{uc}$, $\epsilon_\text{c}$, $\beta_\text{cfg}$
	are the model's final output, unconditional output, conditional
	output, and a user-specified weight respectively.
	When a conditioning image is
	added via ControlNet, it can be added to both $\epsilon_\text{uc}$ and
	$\epsilon_\text{c}$, or only to the $\epsilon_\text{c}$. In
	challenging cases, \eg, when no prompts are given, adding it to both
	$\epsilon_\text{uc}$ and $\epsilon_\text{c}$ will completely remove
	CFG guidance (Figure\,\ref{fig:cfg}b); using only $\epsilon_\text{c}$ will make
	the guidance very strong (Figure\,\ref{fig:cfg}c). Our solution is to
	first add the conditioning image to $\epsilon_\text{c}$ and then multiply a
	weight $w_i$ to each connection between Stable Diffusion and ControlNet according to
	the resolution of each block $w_i=64/h_i$, where $h_i$ is the size of
	$i^\text{th}$ block, \eg, $h_1=8, h_2=16, ..., h_{13}=64$.  By
	reducing the CFG guidance strength , we can achieve the result shown in
	Figure\,\ref{fig:cfg}d, and we call this CFG Resolution Weighting.
	
	\begin{table}[!t]
		\centering
		\resizebox{\linewidth}{!}{\begin{tabular}{@{}lrr@{}}
				\toprule
				Method   & Result Quality $\uparrow$ & Condition Fidelity $\uparrow$ \\  
				\midrule
				PITI~\cite{wang2022pretraining}(sketch)& 1.10 $\pm$ 0.05  & 1.02 $\pm$ 0.01 \\
				Sketch-Guided~\cite{voynov2022sketch} ($\beta=1.6$)& 3.21 $\pm$ 0.62  & 2.31 $\pm$ 0.57 \\
				Sketch-Guided~\cite{voynov2022sketch} ($\beta=3.2$)& 2.52 $\pm$ 0.44  & 3.28 $\pm$ 0.72 \\
				ControlNet-lite& 3.93 $\pm$ 0.59  & 4.09 $\pm$ 0.46 \\
				ControlNet & \textbf{4.22 $\pm$ 0.43}  & \textbf{4.28 $\pm$ 0.45} \\ 
				\bottomrule
		\end{tabular}}
		\vspace{-7pt}
		\caption{Average User Ranking (AUR) of result quality and condition fidelity. We report the user preference ranking (1 to 5 indicates worst to best) of different methods.}
		\vspace{-7pt}
		\label{tab:aur}
	\end{table}
	
	\para{Composing multiple ControlNets.}
	To apply multiple conditioning images (\eg, Canny edges, and pose) to
	a single instance of Stable Diffusion, we can directly add the outputs
	of the corresponding ControlNets to the Stable Diffusion model (Figure\,\ref{fig:multi}). No
	extra weighting or linear interpolation is necessary for such composition.
	
	\begin{figure*}[!t]
		\centering
		\vspace{-12pt}
		\includegraphics[width=\linewidth]{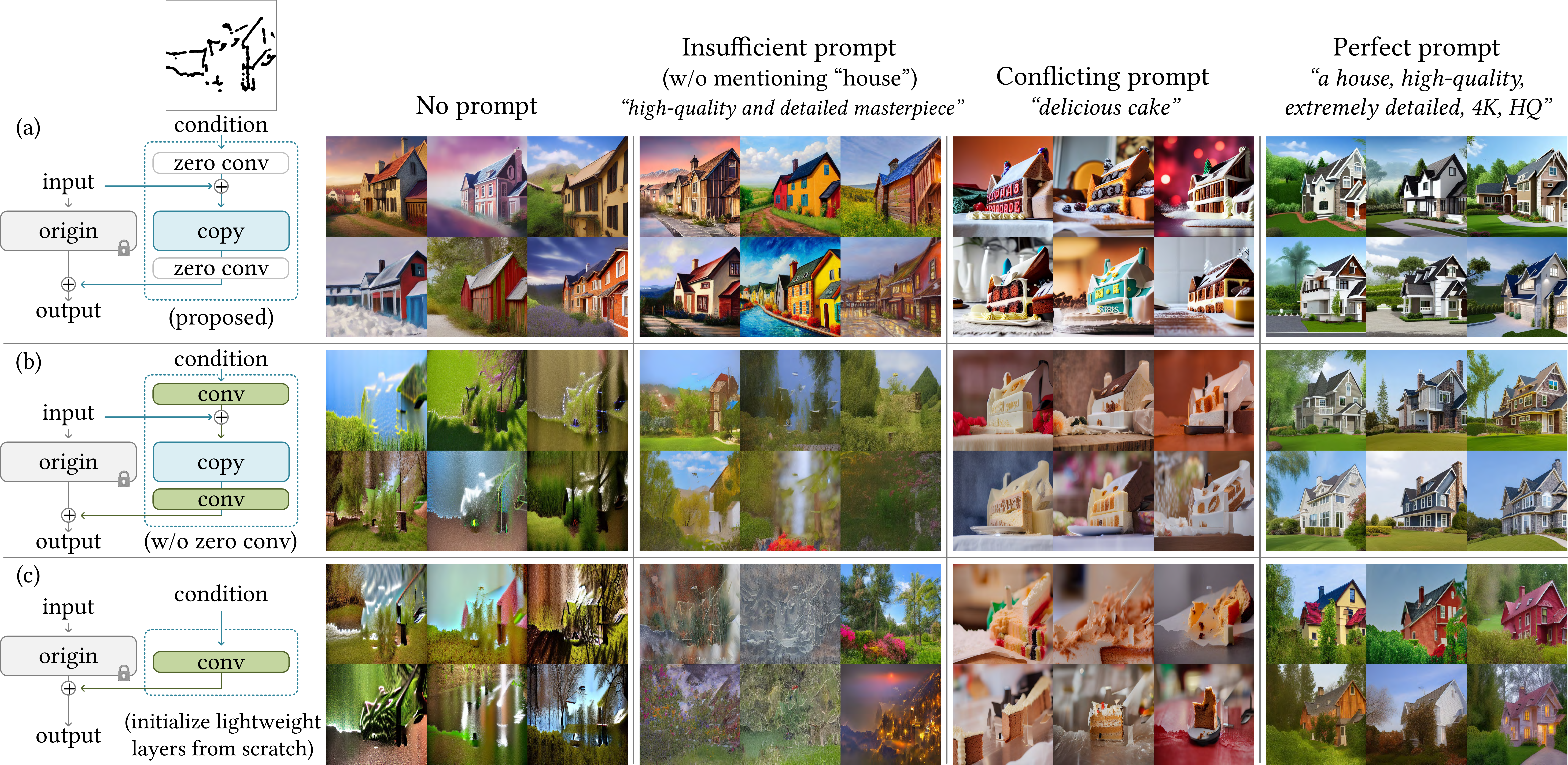}
		\vspace{-18pt}
		\caption{Ablative study of different architectures on a sketch condition and different prompt settings. 
			For each setting, we show a random batch of 6 samples without cherry-picking. Images are at 512 $\times$ 512 and best viewed when zoomed in. The green ``conv'' blocks on the left are standard convolution layers initialized with Gaussian weights.
		}
		\vspace{-10pt}
		\label{fig:abla}
	\end{figure*}
	
	\section{Experiments}
	\label{sec:exp}
	
	We implement ControlNets with Stable Diffusion to test various
	conditions, including Canny Edge\,\cite{canny1986computational}, Depth
	Map\,\cite{ranftl2020towards}, Normal Map\,\cite{diode_dataset}, M-LSD
	lines\,\cite{gu2021realtime}, HED soft
	edge\,\cite{xie2015holistically}, ADE20K
	segmentation\,\cite{zhou2017scene}, Openpose\,\cite{cao2019openpose},
	and user sketches. See also the supplementary material for 
	examples of each conditioning along with detailed training and inference parameters.
	
	\subsection{Qualitative Results}

	Figure\,\ref{fig:tea} shows the generated images in several
	prompt settings. Figure\,\ref{fig:qua} shows our results with
	various conditions without prompts, where the
	ControlNet robustly interprets content semantics in
	diverse input conditioning images.
	
	\subsection{Ablative Study}
	
	\begin{figure}[!t]
		\centering
		\includegraphics[width=\linewidth]{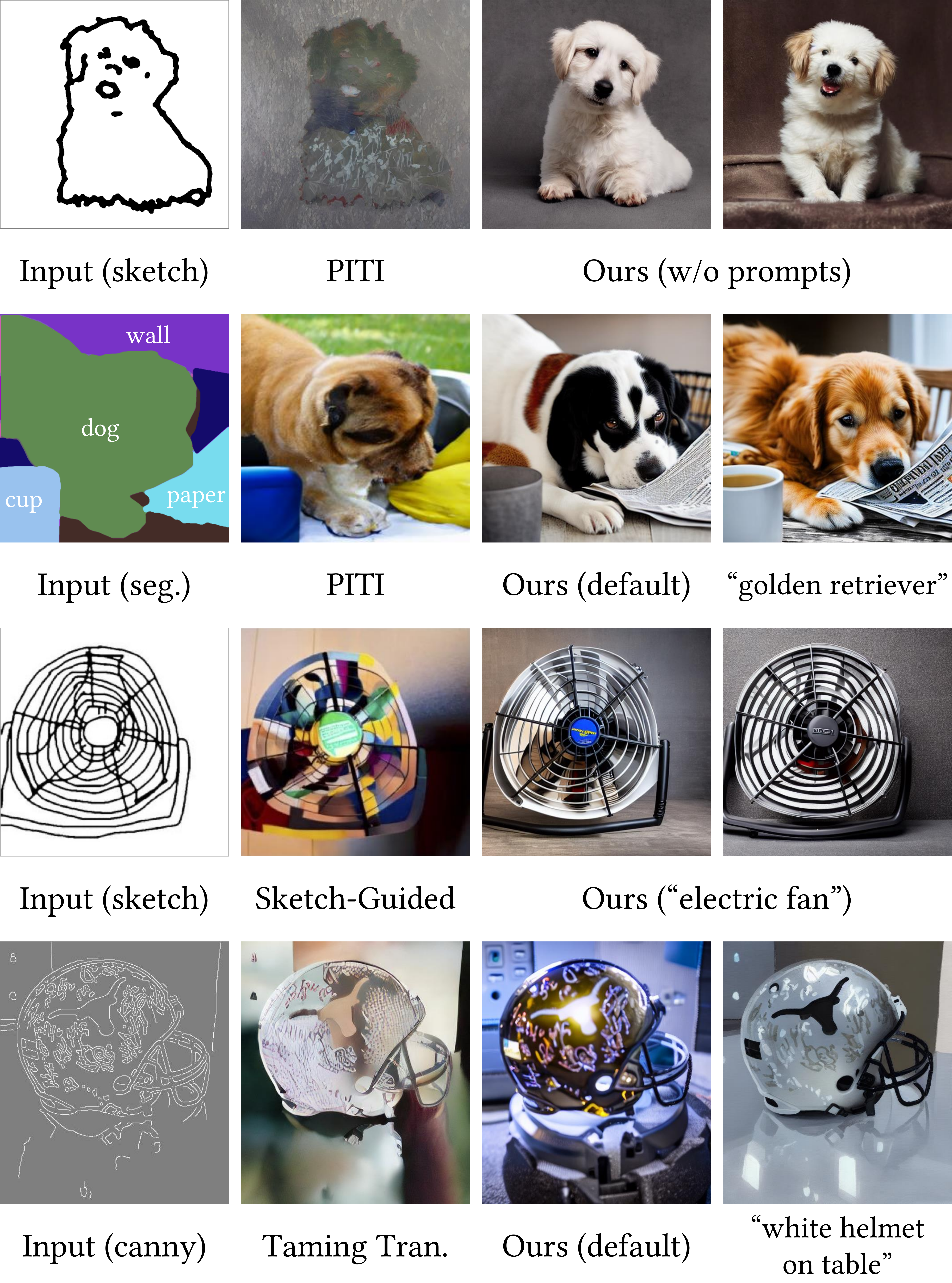}
		\vspace{-20pt}
		\caption{Comparison to previous methods.We present the qualitative comparisons to PITI~\cite{wang2022pretraining}, Sketch-Guided Diffusion~\cite{voynov2022sketch}, and Taming Transformers~\cite{esser2021taming}.}
		\vspace{-10pt}
		\label{fig:compa}
	\end{figure}
	
	We study alternative structures of ControlNets by (1) replacing the zero convolutions with standard convolution layers initialized with Gaussian weights, and (2) replacing each block's trainable copy with one single convolution layer, which we call ControlNet-lite.
	See also the supplementary material for the full details of these ablative structures.
	
	We present 4 prompt settings to test with possible behaviors of real-world users:
	(1) no prompt;
	(2) insufficient prompts that do not fully cover objects in conditioning images, \eg, the default prompt of this paper ``a high-quality, detailed, and professional image'';
	(3) conflicting prompts that change the semantics of conditioning images;
	(4) perfect prompts that describe necessary content semantics, \eg, ``a nice house''.
	Figure\,\ref{fig:abla}a shows that ControlNet succeeds in all 4 settings. 
	The lightweight ControlNet-lite (Figure\,\ref{fig:abla}c) is not strong enough to interpret the conditioning images and fails in the insufficient and no prompt conditions.
	When zero convolutions are replaced, the performance of ControlNet drops to about the same as ControlNet-lite, indicating that the pretrained backbone of the trainable copy is destroyed during finetuning (Figure\,\ref{fig:abla}b).
	
	\begin{table}[!t]
		\centering
		\resizebox{\linewidth}{!}{\begin{tabular}{@{}cccccc@{}}
				\toprule
				ADE20K (GT)   & VQGAN~\cite{esser2021taming} & LDM~\cite{rombach2021highresolution} & PITI~\cite{wang2022pretraining} & ControlNet-lite & ControlNet \\  
				0.58 $\pm$ 0.10 &0.21 $\pm$ 0.15 &0.31 $\pm$ 0.09 &0.26 $\pm$ 0.16 &0.32 $\pm$ 0.12 &\textbf{0.35 $\pm$ 0.14} \\
				\bottomrule
		\end{tabular}}
		\vspace{-8pt}
		\caption{Evaluation of semantic segmentation label reconstruction (ADE20K) with Intersection over Union (IoU $\uparrow$).}
		\vspace{-9pt}
		\label{tab:iou}
	\end{table}

	\subsection{Quantitative Evaluation}
	
	\para{User study.} We sample 20 unseen hand-drawn sketches, and then assign each sketch to 5 methods: 
	PITI\,\cite{wang2022pretraining}'s sketch model,
	Sketch-Guided Diffusion (SGD)\,\cite{voynov2022sketch} with default edge-guidance scale ($\beta=1.6$),
	SGD~\cite{voynov2022sketch} with relatively high edge-guidance scale ($\beta=3.2$),
	the aforementioned ControlNet-lite, and ControlNet. We invited 
	12 users to rank these 20 groups of 5 results individually in terms of \emph{``the quality of displayed images''} and \emph{``the fidelity to the sketch''}.
	In this way, we obtain 100 rankings for result quality and 100 for condition fidelity.
	We use the Average Human Ranking (AHR) as a preference metric where users rank each result on a scale of 1 to 5 (lower is worse).
	The average rankings are shown in Table\,\ref{tab:aur}.
	
	\para{Comparison to industrial models.} Stable Diffusion V2 Depth-to-Image (SDv2-D2I)~\cite{sdd} is trained with a large-scale NVIDIA A100 cluster, thousands of GPU hours, and more than 12M training images. We train a ControlNet for the SD V2 with the same depth conditioning but only use 200k training samples, one single NVIDIA RTX 3090Ti, and 5 days of training. We use 100 images generated by each SDv2-D2I and ControlNet to teach 12 users to distinguish the two methods. Afterwards, we generate 200 images and ask the users to tell which model generated each image. The average precision of the users is $0.52 \pm 0.17$, indicating that the two method yields almost indistinguishable results.
	
	\begin{table}[!t]
		\centering
		\resizebox{0.78\linewidth}{!}{

		\begin{tabular}{@{}lccc@{}}
			\toprule
			Method   & FID $\downarrow$ & CLIP-score $\uparrow$ & CLIP-aes. $\uparrow$ \\  
			\midrule
			Stable Diffusion & 6.09 & 0.26 & 6.32 \\
			\midrule
			VQGAN~\cite{esser2021taming}(seg.)* & 26.28 & 0.17 & 5.14 \\
			LDM~\cite{rombach2021highresolution}(seg.)* & 25.35 & 0.18 & 5.15 \\
			PITI~\cite{wang2022pretraining}(seg.)& 19.74  & 0.20 & 5.77 \\
			ControlNet-lite& 17.92  & 0.26 & 6.30 \\
			ControlNet & 15.27  & 0.26 & 6.31 \\ 
			\bottomrule
		\end{tabular}

}
		\vspace{-7pt}
		\caption{Evaluation for image generation conditioned by semantic segmentation. We report FID, CLIP text-image score, and CLIP aesthetic scores for our method and other baselines. We also report the performance of Stable Diffusion without segmentation conditions. Methods marked with ``*'' are trained from scratch.}
		\vspace{-9pt}
		\label{tab:fid}
	\end{table}
	
	\para{Condition reconstruction and FID score.}
	We use the test set of ADE20K\,\cite{zhou2017scene} to evaluate the conditioning fidelity.
	The state-of-the-art segmentation method OneFormer\,\cite{jain2022oneformer} achieves an Intersection-over-Union (IoU) with 0.58 on the ground-truth set.
	We use different methods to generate images with ADE20K segmentations and then apply OneFormer to detect the segmentations again to compute the reconstructed IoUs (Table\,\ref{tab:iou}).
	Besides, we use Frechet Inception Distance (FID)\,\cite{NIPS2017_8a1d6947} to measure the distribution distance over randomly generated 512$\times$512 image sets using different segmentation-conditioned methods, as well as text-image CLIP scores\,\cite{radford2021learning} and CLIP aesthetic score\,\cite{schuhmann2022laionb} in Table\,\ref{tab:fid}. See also the supplementary material for detailed settings.
	
	\subsection{Comparison to Previous Methods}
	Figure\,\ref{fig:compa} presents a visual comparison of baselines and our method (Stable Diffusion + ControlNet). Specifically, we show the results of PITI\,\cite{wang2022pretraining}, Sketch-Guided Diffusion\,\cite{voynov2022sketch}, and Taming Transformers\,\cite{esser2021taming}. (Note that the backbone of PITI is OpenAI GLIDE\,\cite{glide} that have different visual quality and performance.)
	We observe that ControlNet can robustly handle diverse conditioning images and achieves sharp and clean results.
	
	\begin{figure}[!t]
		\includegraphics[width=\linewidth]{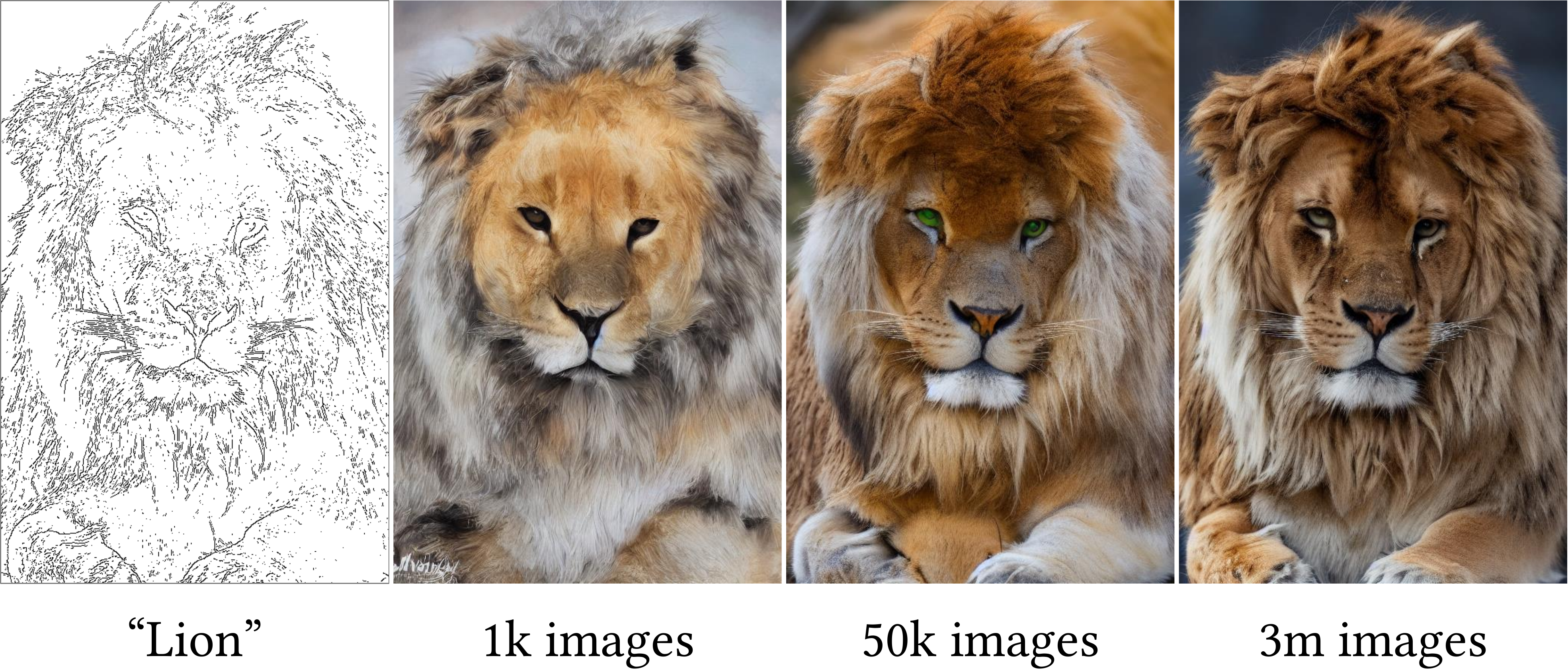}
		\vspace{-18pt}
		\caption{The influence of different training dataset sizes. See also the supplementary material for extended examples.}
		\vspace{-1pt}
		\label{fig:datasize}
		\vspace{-10pt}
	\end{figure}
	
	\begin{figure}
		\includegraphics[width=\linewidth]{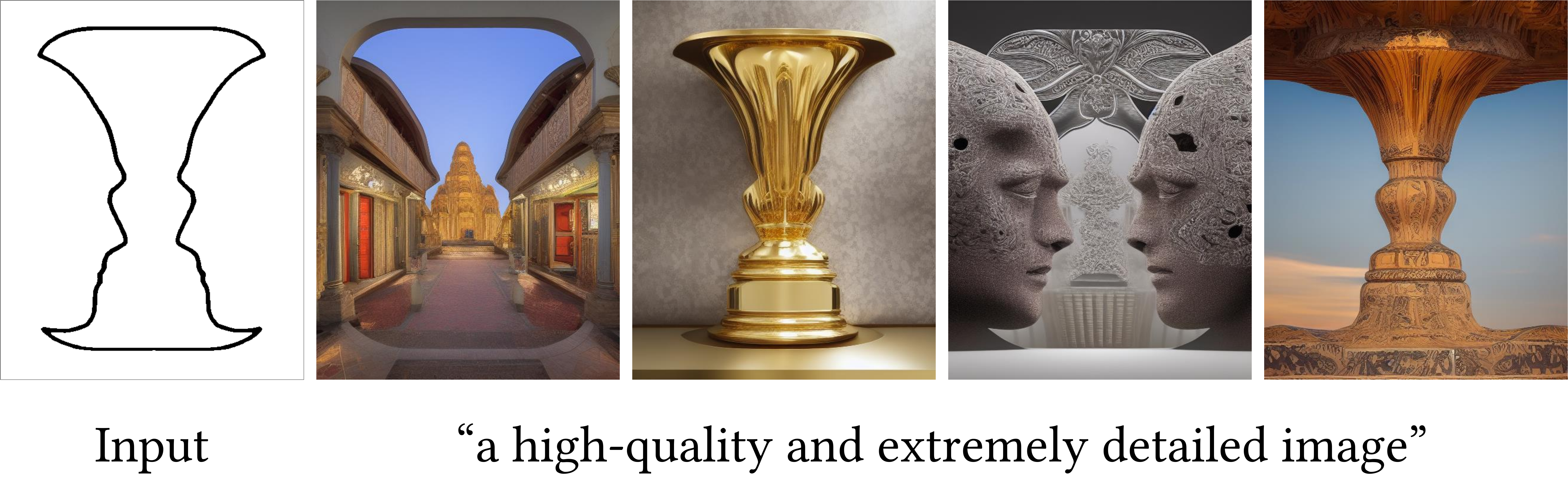}
		\vspace{-18pt}
		\caption{Interpreting contents. If the input is ambiguous and the user does not mention object contents in prompts, the results look like the model tries to interpret input shapes.}
		\vspace{-7pt}
		\label{fig:guess}
	\end{figure}
	
	\begin{figure}[!t]
		\includegraphics[width=\linewidth]{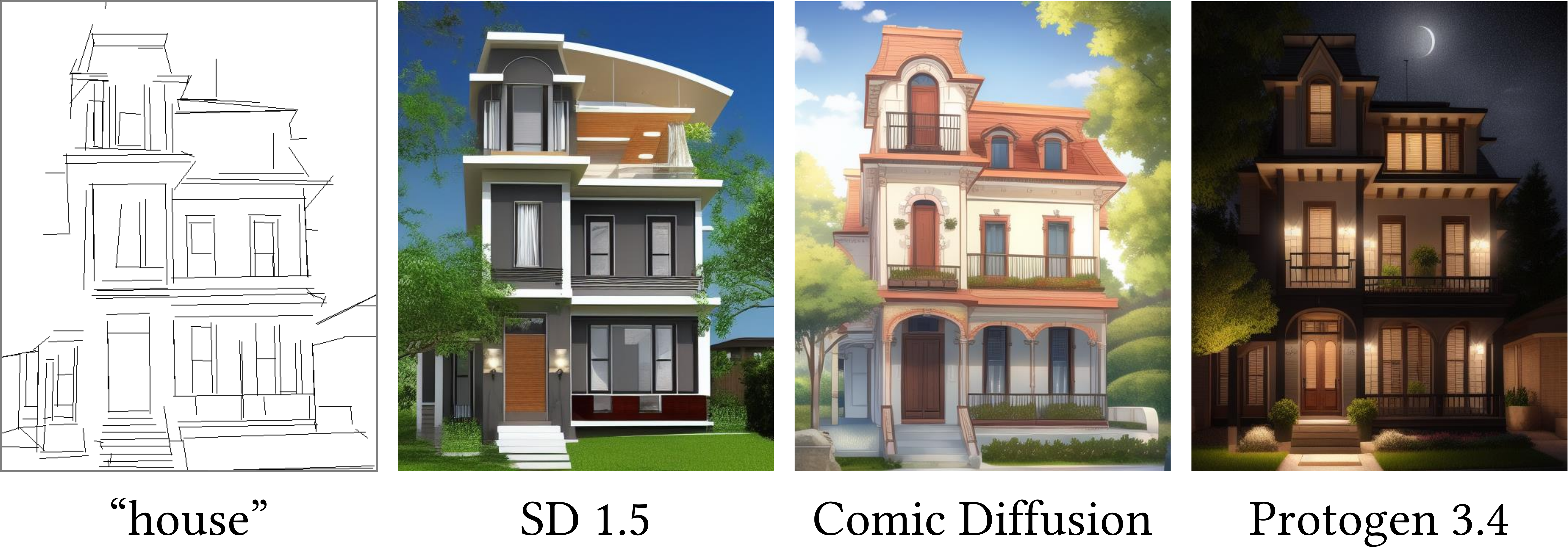}
		\vspace{-18pt}
		\caption{Transfer pretrained ControlNets to community models\,\cite{protogen,comicdiff} without training the neural networks again.}
		\vspace{-1pt}
		\label{fig:trans}
		\vspace{-10pt}
	\end{figure}
	
	\subsection{Discussion}
	
	\para{Influence of training dataset sizes.}
	We demonstrate the robustness of the ControlNet training in Figure\,\ref{fig:datasize}.
	The training does not collapse with limited 1k images, and allows the model to generate a recognizable lion.
	The learning is scalable when more data is provided.
	
	\para{Capability to interpret contents.}
	We showcase ControlNet's capability to capture the semantics from input conditioning images in Figure\,\ref{fig:guess}.
	
	\para{Transferring to community models.}
	Since ControlNets do not change the network topology of pretrained SD models, it can be directly applied to various models in the stable diffusion community, such as Comic Diffusion\,\cite{comicdiff} and Protogen~3.4\,\cite{protogen}, in Figure\,\ref{fig:trans}. 
	
	\section{Conclusion}
	\label{sec:conclusion}
	
	ControlNet is a neural network structure that learns
	conditional control for large pretrained text-to-image
	diffusion models. It reuses the large-scale pretrained layers
	of source models to build a deep and strong encoder to learn
	specific conditions. The original model and trainable copy are
	connected via ``zero convolution'' layers that eliminate
	harmful noise during training.  Extensive experiments verify
	that ControlNet can effectively control Stable Diffusion with
	single or multiple conditions, with or without prompts.
	Results on diverse conditioning datasets show that the
	ControlNet structure is likely to be applicable to a wider
	range of conditions, and facilitate relevant applications.

\section*{Acknowledgment}

This work was partially supported by the Stanford Institute for Human-Centered AI and the Brown Institute for Media Innovation. 

	{\small
		\bibliographystyle{ieee_fullname}
		\bibliography{controlnet}
	}
	
\end{document}